\documentclass[a4paper,conference]{IEEEtran}

\usepackage[pagebackref=false,breaklinks=true,colorlinks,bookmarks=false]{hyperref}

\ifCLASSINFOpdf
  \usepackage[pdftex]{graphicx}
\else
\fi
\usepackage{amsmath}
\usepackage{cleveref}
\usepackage{url}

\usepackage{multirow}
\usepackage[table, dvipsnames]{xcolor}
\usepackage{amsfonts}

\usepackage{soul}
\usepackage{color}

\definecolor{revieweroneColor}{rgb}{1., 1., 0.}
\definecolor{reviewertwoColor}{rgb}{1., .75, .8}
\definecolor{reviewerthreeColor}{rgb}{0., 1., 0.}

\DeclareRobustCommand{\reviewertwo}[1]{{\sethlcolor{reviewertwoColor}\hl{#1}}}

\usepackage{fancyhdr}
\fancypagestyle{firststyle}
{
   \fancyhf{}
   \fancyhead[C]{
    \textcolor{red}{
       The content of this paper was published in ICPR 2022. If you find this research useful, please cite the following paper: \\ Ž. Babnik, P. Peer, V. Štruc, "Faceqan: Face image quality assessment through adversarial noise exploration." in 2022 26th International Conference on Pattern Recognition (ICPR). IEEE Computer Society, 2022. p. 748-754.}
    }
}

\newcommand{\secref}[1]{\S\ref{#1}}

\renewcommand{\baselinestretch}{0.945}
\begin{document}
%
\title{FaceQAN: Face Image Quality Assessment Through Adversarial Noise Exploration}

\author{\IEEEauthorblockN{Žiga Babnik}
\IEEEauthorblockA{Faculty of Electrical Engineering, 
University of Ljubljana\\
Tržaška cesta 25, 1000 Ljubljana\\
ziga.babnik@fe.uni-lj.si}
\and
\IEEEauthorblockN{Vitomir Štruc}
\IEEEauthorblockA{Faculty of Electrical Engineering, 
University of Ljubljana\\
Tržaška cesta 25, 1000 Ljubljana\\
vitomir.struc@fe.uni-lj.si}
}


%
\author{\IEEEauthorblockN{Žiga Babnik\IEEEauthorrefmark{1},
Peter Peer\IEEEauthorrefmark{2} and
Vitomir Štruc\IEEEauthorrefmark{1}}
\IEEEauthorblockA{\IEEEauthorrefmark{1}University of Ljubljana, Faculty of Electrical Engineering
Tržaška cesta 25, 1000 Ljubljana, Slovenia\\}
\IEEEauthorblockA{\IEEEauthorrefmark{2}University of Ljubljana, Faculty of Computer and Information Science, Večna Pot 113, 1000 Ljubljana, Slovenia\\
E-mail: \{ziga.babnik, vitomir.struc\}@fe.uni-lj.si, peter.peer@fri.uni-lj.si}
}


\maketitle
\thispagestyle{firststyle}

\begin{abstract}
Recent state-of-the-art face recognition (FR) approaches have achieved impressive performance, yet unconstrained face recognition still represents an open problem. Face image quality assessment (FIQA) approaches aim to estimate the quality of the input samples that can help provide information on the confidence of the recognition decision and eventually lead to improved results in challenging scenarios. While much progress has been made in face image quality assessment in recent years, computing reliable quality scores for diverse facial images and FR models remains challenging. 
In this paper, we propose a novel approach to face image quality assessment, called FaceQAN, that is 
based on adversarial examples and relies on the analysis of adversarial noise which can be calculated with any FR model learned by using some form of gradient descent. As such, the proposed approach is the first to link image quality to adversarial attacks. 
Comprehensive (cross-model as well as model-specific) experiments are conducted with four benchmark datasets, i.e., LFW, CFP--FP, XQLFW and IJB--C, four FR models, i.e., CosFace, ArcFace, CurricularFace and ElasticFace, and in comparison to seven state-of-the-art FIQA methods to demonstrate the performance of FaceQAN. Experimental results show that FaceQAN achieves competitive results, while exhibiting several desirable characteristics. The source code for FaceQAN is available at \url{https://github.com/LSIbabnikz/FaceQAN}. 
\end{abstract}


%
\IEEEpeerreviewmaketitle

\section{Introduction}\label{introduction}
In recent years, Face Recognition~(FR) techniques have achieved excellent results on datasets containing both high quality frontal images and images of variable quality \cite{grm2018strengths,grm2018deep,meden2021privacy}. Face recognition in completely unconstrained settings, on the other hand, remains challenging, as no quality guarantees can be made for facial images captured in-the-wild~\cite{wang2021deep,MFR_IJCB2021,grm2019face}. 
Face Image Quality Assessment~(FIQA) methods aim to improve the face recognition performance in such settings, by providing additional information to the FR models corresponding to the quality of the input face images.  
Based on this information, 
a FR model can reject low quality images, which often times cause critical False Non-Match (FNM) errors \cite{surveypaper}. 

According to ISO/IEC 29794-1, the quality of a biometric sample can be defined using its character, fidelity or utility~\cite{isoiec}. Contemporary FIQA methods typically rely on the latter 
and most often define quality in terms of the utility of the input face sample for the FR task. Here, the term \textit{utility} is used to describe the fitness of a sample to accomplish or fulfill a
biometric function \cite{surveypaper}. In this setting, a single numerical score is commonly calculated to capture the quality of the given face image \cite{surveypaper}. From  a conceptual view point, contemporary state-of-the-art FIQA techniques can be broadly partitioned into two distinct groups: i.e., $(i)$ regression-based methods and $(ii)$ model-based approaches. Techniques from the first group learn a mapping directly from the image space to automatically-generated pseudo ground-truth quality labels. Several labeling approaches have been introduced over the years to facilitate this process, the basis for which are comparison/similarity scores calculated using a selected FR model. Recent techniques, for example, consider either comparison-score distributions generated over mated image pairs from a given dataset as the basis for the reference quality labels~\cite{pcnet, lightqnet}, or alternatively, the distribution of similarity scores between probe samples and mated reference samples of a known quality~\cite{faceqnet}. Techniques from the second group, on the other hand, integrate the quality estimation process directly with the considered FR model and commonly predict the quality of the input samples based on the uncertainty of the computed face representations (features/embeddings)~\cite{pfe, magface, serfiq}. Such techniques are tightly linked to the given FR model and usually designed without the need for supervised learning. While the quality scores produced by existing approaches from either group have been shown to be good predictors of the utility of the input images for face recognition, they are still limited by the validity of the pseudo ground truth labels and poor integration with the targeted FR model. 

\begin{figure}[!t]
    \begin{minipage}{0.02\columnwidth}
    \ \
    \end{minipage}
    \begin{minipage}{0.31\columnwidth}
    \centering
    \text{\scriptsize \ \ High quality} \vspace{1mm}
    \end{minipage}
    \begin{minipage}{0.32\columnwidth}
    \centering
    \text{\scriptsize \ \ Medium quality} \vspace{1mm}
    \end{minipage}
    \begin{minipage}{0.32\columnwidth}
    \centering
    \text{\scriptsize \ \ \ Low quality } \vspace{1mm}
    \end{minipage}
    \begin{minipage}{0.02\columnwidth}
    \rotatebox{90}{\scriptsize \ \ Noise   \ \ \ Sample   }
    \end{minipage}
    \begin{minipage}{0.965\columnwidth}
    \includegraphics[width=\linewidth, trim = 5mm 0 0 14mm, clip]
    {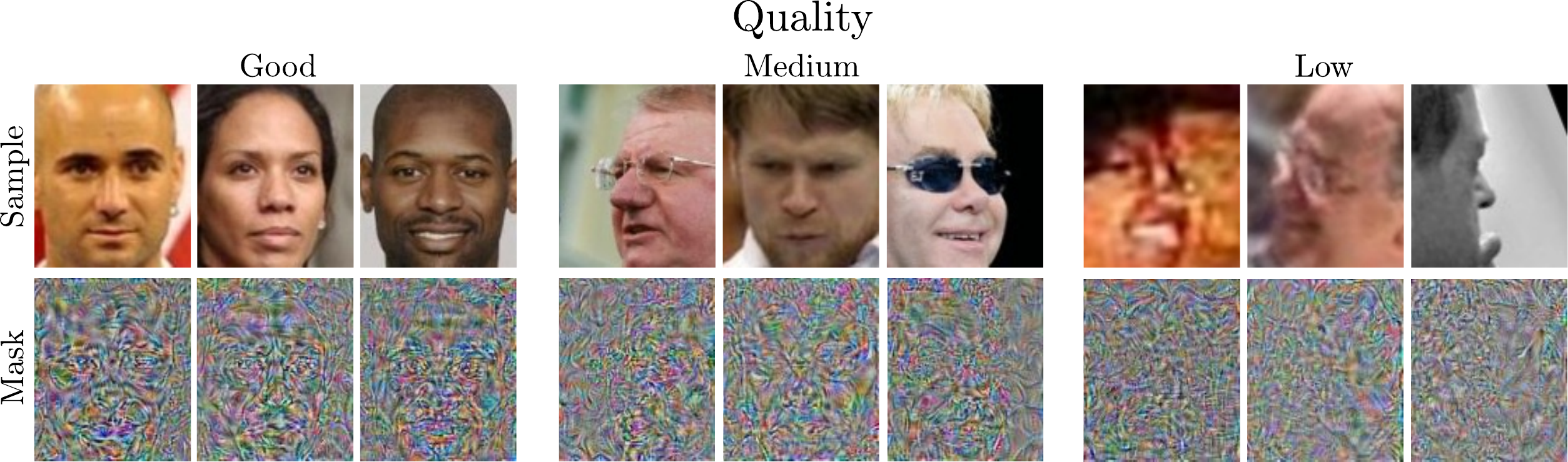}
    \end{minipage}
    \caption{\textbf{Visualization of adversarial noise as a function of image quality.} 
    A clear difference can be observed in the characteristics of the adversarial noise between  images of different quality. Higher quality images produce noise masks with a more distinct face-like pattern, while the lower quality masks are closer to random noise. FaceQAN takes advantage of the presented observations to estimate the quality of the input images.\vspace{-2mm}}
    \label{fig:intro_img}
\end{figure}

In this paper, we propose a novel, unsupervised FIQA approach called FaceQAN~(Face Image Quality Assessment Through Adversarial Noise Exploration) that 
avoids the pseudo quality label generation process and, consequently, the supervised learning process, by using information available in the input facial image and a given (targeted) FR model only. The basis for FaceQAN are adversarial examples, which can be generated for all modern FR models learned through some form of gradient descent \cite{akhtar2021advances}. The proposed method works under the assumption that the difficulty of adversarial example generation is directly correlated with the quality of a given image w.r.t. a certain FR model, as shown on Fig~\ref{fig:intro_img}. In other words, good quality images are expected to produce stable and robust representations that are difficult to perturb using adversarial noise. Based on this insight, we make the following main contributions in this paper:
\begin{itemize}
    \item We propose FaceQAN, a novel approach for generating quality scores for face images that ensures competitive results on several datasets and in comparison to a wide range of recent state-of-the-art FIQA techniques. 
    \item To the best of our knowledge, we are the first to link adversarial attacks to the task of face image quality assessment and show that such linkage leads to highly desirable FIQA characteristics.
\end{itemize}

\section{Related Work}\label{related_work}

In this section, we 
position FaceQAN with respect to the two main groups of existing methods, i.e., regression- and model-based approaches. 
A more comprehensive coverage of the field can be found in the recent survey paper in
\cite{surveypaper}.

\textbf{Regression-Based FIQA.} 
Techniques from this group utilize pseudo quality (ground truth) labels for the prediction of quality scores. The ground-truth label-generation process is usually automatic~\cite{faceqnet, sddfiqa, lightqnet} but can also involve human assessment~\cite{mix1, mix2}. The most recent methods in the literature utilize deep learning models paired with automatic extraction of ground truth labels. One such method, called \textit{FaceQNet}, presented by Hernandez-Ortega \textit{et al.} in~\cite{faceqnet}, is based on a ResNet-$50$ model trained on pseudo labels generated from the VGGFace2~\cite{vggface2} dataset. The ground truth labels are obtained by first selecting the highest quality image of the individuals in the set, calculated using third party ISO/IEC 19794-5~\cite{biolab-icao, isoiec2} compliance software. For all other images in the set, the quality is calculated as the normalized Euclidean distance between the embeddings of the given image and of the highest quality image of a particular individual. A pretrained ResNet-$50$ network is then fine-tuned on the generated ground truth labels. A more advanced labeling process is presented by Ou \textit{et al.}~\cite{sddfiqa} in the \textit{SDD-FIQA} method. Here, the authors suggest using both the inter-class and intra-class distances to produce ground truth labels. For a given image, both mated and non-mated comparison-score distributions are constructed first. 
The quality score is then computed as the average Wasserstein distance between the distributions over several runs. A face recognition model is finally fine-tuned on the obtained ground truth labels, similar to FaceQNet. 

While the discussed approaches perform well on standard datasets such as LFW~\cite{lfw}, they heavily rely on the construction of ground truth quality labels. This process often introduces biases associated with either the utilized FR model or the selected dataset. Since the end goal is to improve the face recognition performance of a particular FR model, with its own associated biases, such FIQA methods may not produce optimal quality scores for different datasets and FR models. To avoid such shortcomings, the proposed FaceQAN quality estimation approach  relies only on the information contained in the given image and a targeted FR model, avoiding any learning and, consequently, the process of generating pseudo ground truth labels. In cases where the targeted FR model is not available, the FaceQAN quality assessment process can also be performed with an arbitrary surrogate FR model.

\textbf{Model-Based FIQA.} Solutions in this group typically combine the face recognition and FIQA tasks, and learn to produce the embeddings and the quality (or uncertainty) of the input image simultaneously. One such method, \textit{PFE} (Probabilistic Face Embeddings) presented by Shi and Jain~\cite{pfe}, learns to predict the uncertainty of a given image, by estimating the 
mean and variance of the computed feature vectors. Here, the mean vector corresponds to the embedding of the given image, whereas the variance presents the uncertainty of the image in the feature space. A quality score is obtained by calculating the harmonic mean over the variance vector. More recently Meng \textit{et al.}~\cite{magface} presented \textit{MagFace} an approach that incorporates the prediction of the image uncertainty into the face-recognition learning process. The authors achieved this by introducing a new loss function built upon the ArcFace loss, which enables better separation for embeddings with higher magnitudes. The quality score of a given image is calculated simply as the norm of its embeddings, if the embedding model is trained using the presented loss. While such methods do not need to explicitly extract quality ground truth labels, they enforce a learning regime upon the targeted FR model. The proposed FaceQAN method, on the other hand, is \textit{learning-free} and applicable directly to any (pretrained) targeted model \cite{akhtar2021advances} without the need for interventions into the learning procedure.


Our approach is most closely related to \textit{SER-FIQ}, presented by Terhörst \textit{et al.} in~\cite{serfiq}. SER-FIQ uses the properties of \textit{dropout}, a regularization technique commonly used in modern convolutional neural networks (CNNs) to avoid overfitting. More precisely, several embeddings are first produced for a single input image using different sub-network layouts, generated using dropout. The quality score is then computed as the sigmoid of the negative normalized sum of distances over all pairs of produced embeddings. Similarly to SER-FIQ, FaceQAN also aims to capture the uncertainty of the embeddings (computed for an input face image) for quality assessment, but does so through the analysis of a set of generated adversarial examples.


\section{Methodology}\label{methodology}

An ideal FIQA method should reflect the biases and performance of the targeted FR model, while being generally applicable to different model topologies trained with arbitrary learning objectives. Thus, only information originating from the targeted FR model and the given input sample should be considered in the quality estimation process. The main contribution of this work, FaceQAN, presented in this section, follows the outlined logic by estimating the sample quality using adversarial noise, which can be generated for all modern FR networks, trained using some variant of gradient descent \cite{akhtar2021advances}, as well as information available directly in the input image. Different from most adversarial methods, our goal is not to create a specific adversarial example, but rather to measure the difficulty of adversarial example generation. 

\subsection{Overview of FaceQAN}\label{methodology:anfiqa}

\begin{figure}
    \centering
    \includegraphics[width=0.99\linewidth]{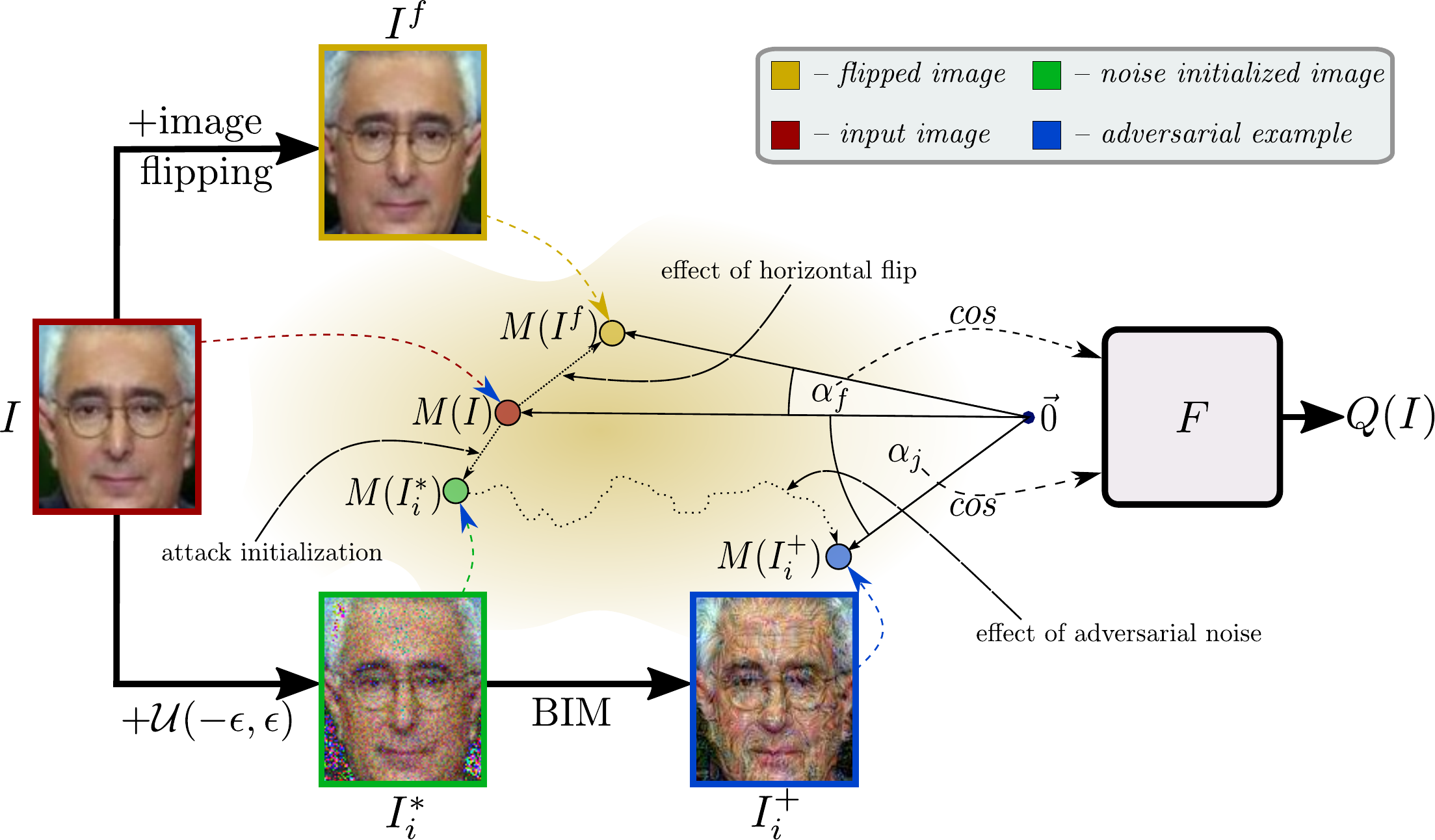}
        \caption{\textbf{High-level overview of FaceQAN.} 
        FaceQAN estimates the quality of an input sample $I$ by exploring the characteristics of adversarial examples $I^+$ in the embedding space of the targeted FR model $M$. The final quality is estimated through an aggregation function $F$ that considers the similarity between the embedding of the input sample and $k$ adversarial examples, i.e., $i=1,\ldots, k$. Moreover, the impact of pose on quality is modeled explicitly within FaceQAN through a face-symmetry estimation procedure based on image flipping. The figure is best viewed electronically and in color.\vspace{-2mm}%
        }
    \label{fig:method_deep}
\end{figure}

Given a targeted FR model $M$ and an input face image $I\in\mathbb{R}^{w\times h\times3}$, the goal of face image quality assessment is to generate a quality score $Q(I)\in\mathbb{R}$ that reflects the utility of $I$ with respect to $M$. With FaceQAN, the quality score for an input image $I$ is computed by analyzing the characteristics of adversarial examples $I^+$ in the embedding space of the targeted FR models, i.e., $Q(M(I),M(I^+))$. 
While the pixel space could also be considered for the exploration of the adversarial noise, using the embedding space ensures tighter integration with the targeted FR models and allows FaceQAN to better capture the utility of the input face samples for FR. 

The complete FaceQAN pipeline, illustrated in Fig.~\ref{fig:method_deep}, consists of four distinct steps: $(i)$ an \textit{attack initialization} step (\secref{methodology:anfiqa:first}) that generates multiple noisy images from the given input face to facilitate the attack procedure, $(ii)$ an \textit{adversarial-example generation} step (\secref{methodology:anfiqa:second}) that produces the attack samples needed for the analysis, $(iii)$ a \textit{symmetry estimation} step (\secref{methodology:anfiqa:third}) that specifically accounts for the impact of pose on image quality, and $(iv)$ a \textit{quality-score calculation} step (\secref{methodology:anfiqa:fourth}) that computes the final quality value based on the analysis of the generated adversarial embeddings.
All steps are described in-detail in the following sections. 

\subsection{Attack Initialization}\label{methodology:anfiqa:first}

Standard adversarial attacks 
typically require an input sample $I$, a FR model $M$ and the ground truth class/identity label 
$y$ of the input sample to be able to generate adversarial examples \cite{fgsm, madry2017towards, carlini2017towards}. Because face image quality assessment  needs to be applicable to arbitrary facial images irrespective of the identities/labels used to learn the model $M$, 
we utilize the similarity-based adversarial attack (SAA) criterion, proposed recently by Wang~\textit{et al.} in
\cite{fgsmextend}, as the basis for FaceQAN and adapt it to fit our problem. SAA allows us to use the embedding of the input image $I$ as the ground truth label to attack, i.e., $y=M(I)$, and to define an (angular) dissimilarity loss to drive the adversarial noise generation process, i.e.:
\begin{equation}
    L(M(I^*),y) = 1-\frac{M(I^*)^T\cdot y}{\|M(I^*)\|\|y\|},  
    \label{eq:cosine_embd}
\end{equation}
where $\|\cdot\|$ is the $L_2$ norm. Since adversarial examples are generated by maximizing $L$, we ensure that the gradient of $L$ with respect to the input $I^*$ is non-zero, and, therefore, define $I^*$ as a noise perturbed version of the original input image $I$. As long as the noise infused into $I$ is minute, the embedding of $I^*$ is expected to be close to the embedding of the original input image $I$, as also illustrated in Fig.~\ref{fig:method_deep}.

Let $I\in\mathbb{R}^{w\times h\times3}$ be an input face image with values in the range of $[-1, 1]$. FaceQAN generates the noisy counterpart to $I$ required for the attack initialization as:
\begin{equation}
    I^* = \lfloor I + N\rfloor_{[-1, 1]}, \;\;\; N \sim \mathcal{U}(-\epsilon, \epsilon),
\end{equation}
where the noise $N$ is sampled from a uniform distribution $\mathcal{U}$, $\epsilon$ is an open hyperparameter, and $\lfloor \cdot \rfloor_{[-1, 1]}$ denotes a clipping operator that guarantees that the resulting noisy image $I^*$ contains pixels in the range $[-1, 1]$. Choosing a 
value of $\epsilon$ close to 0 assures that the amount of noise added to the image is minuscule and that the change in the embedding is limited. 

Because the added noise initializes the adversarial attack in an arbitrary direction within the embedding space, we minimize this randomness by generating  
a set of $k$ noisy images, i.e., $\{I^*_i\}_{i = 1}^k$, by sampling the noise independently from $\mathcal{U}$ $k$ times. This procedure allows FaceQAN to explore the embedding space around $M(I)$ in various directions and estimate the stability of the embedding for quality estimation. 



\subsection{Adversarial Example Generation}\label{methodology:anfiqa:second}

In the second step, FaceQAN uses the set of noisy images $\{I^*_i\}_{i = 1}^k$ to generate a corresponding set of adversarial examples $\{{I}^+_i\}_{i = 1}^k$.  We note again, that the use of adversarial methods in FaceQAN deviates from the usual, as we are not interested in generating adversarial examples that can fool a targeted FR model $M$, but rather in the characteristics of the generated examples  after an attack with \textit{fixed attack hyperparameters}. 
FaceQAN is in general applicable with any adversarial approach, but we select the Fast Gradient Sign Method (FGSM)~\cite{fgsm} together with the Basic Iterative Method~(BIM)~\cite{bim} for the implementation in this paper due to their simplicity and ease of implementation.

\textbf{Fast Gradient Sign Method.} In our setting, FGSM generates an adversarial example $I^+$ from $I$ such that the difference between the true label $y=M(I)$ and the embedding of the adversarial example $M(I^+)$ is maximized. 
The procedure starts by first generating the embedding of the input image $M(I)$ and then calculating 
the loss from Eq.~\eqref{eq:cosine_embd} between the generated prediction and the true label, i.e., $L(M(I), y)$. The calculated loss is then back-propagated to the input $I$, for which the gradient $\nabla_I L(M(I), y)$ is computed. Finally, the adversarial example is constructed by defining the adversarial noise $\mu$ as the sign function of the gradient, i.e., $sign(\nabla_I L(M(I), y)$, and in turn:
    $I^+ = I + \epsilon \cdot \mu$,
where $\epsilon$ is again an open hyperparameter of the method, 
that controls the amount of noise added to $I$. A smaller $\epsilon$ corresponds to a lower amount of adversarial noise in the final adversarial example $I^+$.

\textbf{Basic Iterative Method.} We use BIM to improve the attack capabilities of FGSM and to better explore the embedding space around $M(I)$. BIM extends FGSM by creating adversarial examples over $l$ FGSM runs, where only the input $I$ changes between iterations. Additionally, the parameter $\epsilon$ is scaled as $\frac{\epsilon}{l}$, to limit the overall amount of noise added to the image. The initial iteration of BIM is identical to FGSM, whereas for all consequent iterations $m \in [2, l]$ the input $I$ is defined as the adversarial example $I^{+}_{m-1}$ produced by the previous iteration. Performing $l$ iterations, we obtain the final adversarial example, i.e.:
 $   I^{+} \xleftarrow[]{} BIM_{l}(I, y, M).$
The set of adversarial examples $\{{I}^+_i\}_{i = 1}^k$ is generated from the set of noisy images using simple batch processing.

\subsection{Symmetry Estimation}\label{methodology:anfiqa:third}

A critical factor known to affect face recognition performance are pose variations~\cite{vstruc2010complete,zhao2018towards,banerjee2018frontalize}. For FaceQAN, we, therefore, design an additional step to explicitly account for the effect of pose in the input images. 
To avoid unnecessary complexity, we use the following simple logic. Generally faces are vertically symmetrical and as such, if presented with a fully frontal face image, flipping the image horizontally should not have major effects on the produced embedding. This does not hold true for larger pose variations. Thus, we create a horizontally flipped image $I^f$ for each given input $I$ and use its embedding alongside the embeddings of the adversarial set $\{{I}^+_i\}_{i = 1}^k$ when computing the final quality score.

\subsection{Quality Score Calculation}\label{methodology:anfiqa:fourth}

In the last step, the adversarial set $\{{I}^+_i\}_{i = 1}^k$ and the flipped image $I^f$ are passed through the model $M$ to obtain a set of adversarial embeddings $\{{y}^+_i\}_{i=1}^k = \{\,M(I^{+}_i) \mid i \in [1, k] \,\}$ and flipped embedding $y_f = M(I^f)$. Both are compared to the embedding $y = M(I)$, using the cosine similarity, which well matches the learning objectives of modern FR models:
\begin{equation}
    \cos \alpha_i = S_i =  \frac{y^T\cdot y^+_i  }{\|y\| \|y^+_i\|}, \;\;\; \cos \alpha_f = s_f  = \frac{y^T \cdot y_f}{\|y\| \|y_f\|},
    \label{eq:pose}
\end{equation}
where the set $\{S_i\}_{i=1}^k$ contains similarities for each adversarial example in $\{{I}^+_i\}_{i = 1}^k$, and $\alpha_i$ and $\alpha_f$ are the angles between the embeddings - see Fig.~\ref{fig:method_deep}. The set of similarities $\{S_i\}_{i=1}^k$ and $s_f$ are then used to calculate the final quality score using the aggregation function $F: \{\{S_i\}_{i=1}^k,s_f\} \mapsto Q$,  defined by: 
\begin{equation}
    q_{adv} = \frac{\mu_S + 1}{2} \cdot \lfloor(1 - \sigma_S)\rfloor_{[0, 1]}, \ Q =
    \left(\ q_{adv} \cdot s_f \right)^p
\end{equation}
where $\mu_S$ represents the mean over $\{S_i\}_{i=1}^k$ and $\sigma_S$ is the corresponding standard deviation. Hence, $F$ considers the mean of the similarity scores as well as their dispersion when computing the quality score, and additionally weights the estimated quality with the computed symmetry prediction. The main intuition behind the adversarial part of this procedure is that lower quality images map to a poorly defined part of the embedding space that is easily perturbed, i.e., the adversarial examples have a low average and high standard deviation.

Because the cosine similarity is bound between $[-1, 1]$, we re-scale both the mean and standard deviation to $[0, 1]$. The final quality is as such defined on $[0, 1]$, where higher values represent images of better quality with respect to $M$. To ensure the final quality scores $Q(I)$ cover the full range of values on $[0, 1]$, we use a power law as the last computational step in FaceQAN, where $p$ is an open hyperparameter. Note that this final step has no impact on the quality-estimation process and only affects the range of the aggregation function $F$.

\section{Experiments and Results}\label{evaluation}

\begin{table}[!t!]
    \centering
    \renewcommand{\arraystretch}{1.1}
    \caption{Summary of the experimental setup}\vspace{-2mm}
\resizebox{\columnwidth}{!}{%
    \begin{tabular}{|l|l|c||c|c||c|c|c|c|}
        \hline
        \multirow{ 2}{*}{\textbf{Dataset}} & \multirow{ 2}{*}{\textbf{\#Images}} & \multirow{ 2}{*}{\textbf{\#IDs}} & \multicolumn{2}{c||}{\textbf{\#Comparisons}} & \multicolumn{4}{c|}{\textbf{Main Quality Factors}$^\dagger$$^\ddagger$}\\\cline{4-9}
        
        & & & Mated & Non-mated & Pose & O-E & B-R-N & Sc\\ 
        \hline
        LFW~\cite{lfw} & $13{\small,}233$ & $5{\small,}749$ & $3{\small,}000$ & $3{\small,}000$ & L & L & L & M\\
        CFP-FP~\cite{cfp-fp} & $7{\small,}000$ & $500$ & $3{\small,}500$ & $3{\small,}500$ & H & M & L & M  \\
        XQLFW~\cite{xqlfw} & $13{\small,}233$ & $5{\small,}749$ & $3{\small,}000$ & $3{\small,}000$ & L & L & H & M\\
        IJB-C~\cite{ijbc} & $23{\small,}124^{\dagger\dagger}$ & $3{\small,}531$ & $19{\small,}557$ & $15{\small,}638{\small,}932$ & H & H & H & Lr\\
        \hline
        \multicolumn{8}{l}{\footnotesize $^\dagger$O-E - Occlusion, Expression; B-R-N - Blur, Resolution, Noise; Sc - Scale.}\\
        \multicolumn{8}{l}{\footnotesize $^\ddagger$L - Low; M - Medium; H - High; Lr - Large; Values estimated subjectively by the authors.}\\
        \multicolumn{8}{l}{\footnotesize $^{\dagger\dagger}$ number of templates, each containing several images}
        \vspace{-4mm}
    \end{tabular}
    }
    \label{tab:my_flips}
\end{table}

\begin{figure*}[!t]
    \centering
        \includegraphics[width=0.91\textwidth, trim = 3mm 1mm 3mm 0mm,clip]{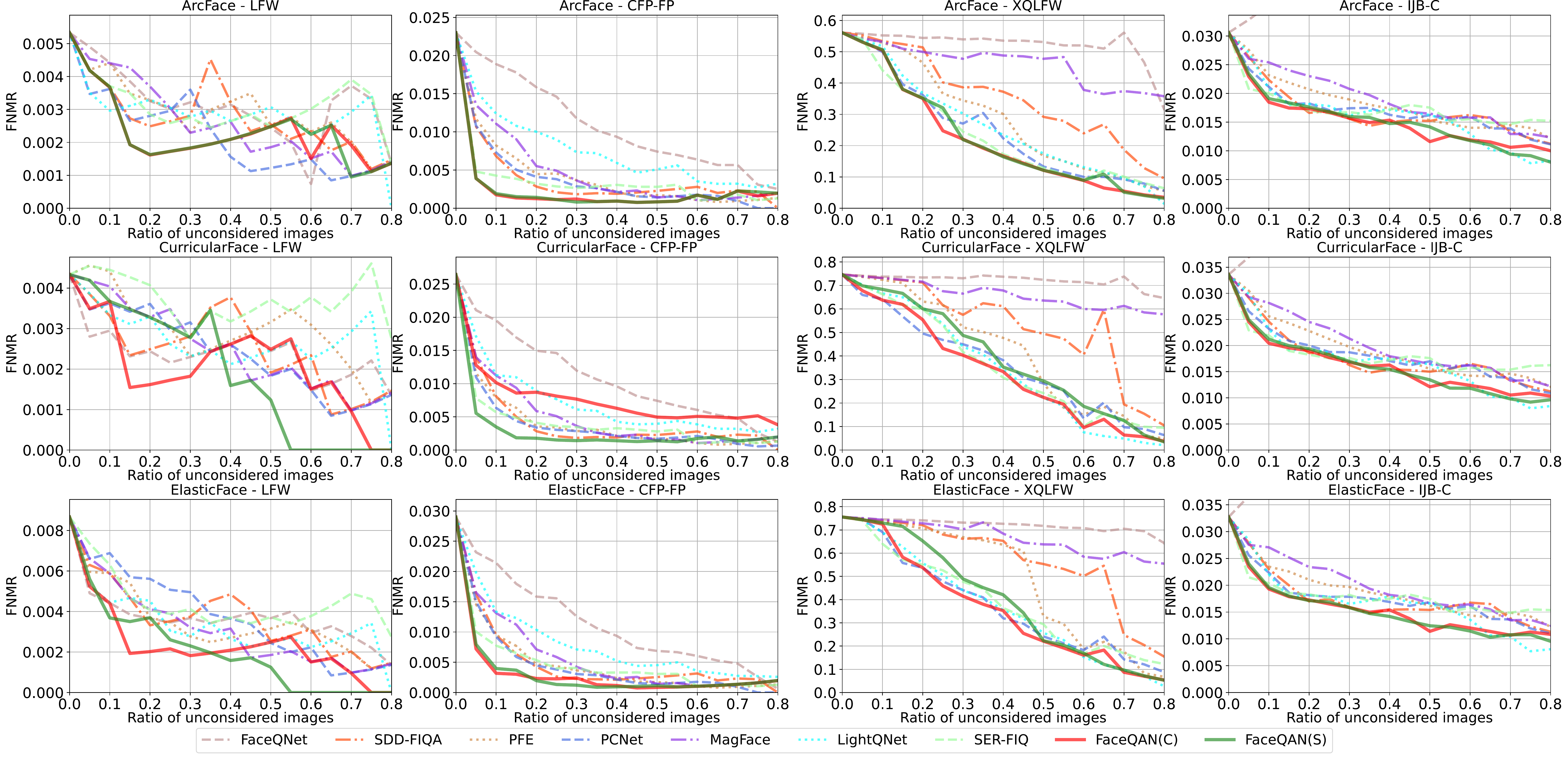}\vspace{-3mm}
    \caption{\textbf{ERC results at FMR=$0.001$}. FaceQAN is evaluated using three state-of-the-art FR models over four datasets and in comparison to seven state-of-the-art baselines and yields highly competitive results. All results correspond to cross-model (C) quality assessment experiments. For reference purposes, model-specific FaceQAN results, marked (S), are also reported. The figure is best viewed electronically and in color.
     \vspace{-1mm}
    }
    \label{fig:erc_res}
\end{figure*}

\subsection{Experimental Setup}

\textbf{Evaluation Setting.} We compare FaceQAN with seven state-of-the-art FIQA methods. We chose FaceQNet~\cite{faceqnet}, MagFace~\cite{magface}, PCNet~\cite{pcnet}, LightQNet~\cite{lightqnet}, and SDD-FIQA~\cite{sddfiqa}, as representatives of regression-based techniques, and PFE~\cite{pfe} and SER-FIQ~\cite{serfiq} as examples of model-based approaches. To ensure a fair comparison, publicly available official implementations are used or code obtained directly from the authors.  As summarized in Table~\ref{tab:my_flips}, we test FaceQAN on four benchmark datasets with diverse (quality) characteristics, i.e.: Labeled Faces in the Wild (LFW)~\cite{lfw}, 
Celebrities in Frontal-Profile in the Wild (CFP-FP)~\cite{cfp-fp}, Cross-Quality LFW (XQLFW)~\cite{xqlfw} and IJB-C (IARPA Janus Benchmark -- C)~\cite{ijbc}, similarly to \cite{boutros2021cr}. 
The experiments are conducted with three popular (state-of-the-art) open-source FR models, referred to as  ArcFace\footnote{\label{note1}\url{https://github.com/deepinsight/insightface}} \cite{arcface}, CurricularFace\footnote{\url{https://github.com/HuangYG123/CurricularFace}} \cite{curricularface}, and ElasticFace\footnote{\url{https://github.com/fdbtrs/ElasticFace}} \cite{elasticface} hereafter based on their learning objectives. 
All three FR models are based on the ResNet100 architecture. ArcFace is trained on MS1MV3, ElasticFace on MS1MV2, whereas CurricularFace is trained on CASIA-WebFace and MS1MV2. 
To enable a fair comparison with the competing state-of-the-art FIQA methods, all results are generated using a cross-model (C) setting, where quality scores are generated using one FR model and performance is measured on another. 
To compute quality scores for FaceQAN we use the CosFace\textsuperscript{\labelcref{note1}} \cite{cosface} model trained with the cosface objective with a ResNet100 architecture on the Glint360K dataset \cite{an2021partial}. For reference purposes, we also provide model-specific FaceQAN results (S), where access to a target FR model is assumed and, hence, the same FR model is used for quality prediction and performance scoring. 


\textbf{Performance Evaluation.} Following standard methodology~\cite{magface,serfiq,sddfiqa,surveypaper}, the Error Versus Reject Characteristic~(ERC) is used to score performance. ERC measures the False Non Match Rate~(FNMR) at a constant False Match Rate~(FMR) usually set to $0.001$, when increasing the fraction of unconsidered images. Images are rejected based on their calculated quality label. Additionally, we also measure the Area Under the Curve (AUC) for the ERC plots, where lower scores imply better performance. As suggested in~\cite{surveypaper}, we report AUC values at several  drop rates, i.e., $0.1$, $0.2$, $0.4$ and $0.8$. Here, smaller drop rates are typically more relevant, since this is when the images of the lowest quality are removed.

\textbf{Implementation Details.} Since we are interested only in exploring the localized space around the computed face embedding, $M(I)$, a combination of a small $\epsilon$ and low $l$ is chosen for the experiments. Specifically, we set $\epsilon=0.001$ and $l=5$ as the values for the BIM parameters. Ideally, the batch size should follow $k \gg d_{M(I)}$, where $d_{M(I)}$ represents the dimensions of the embedding space, so that all directions around $M(I)$ are well explored. However, because the dimensionality of the embedding spaces is usually large ($>512$), we set $k=10$ to ensure a reasonable trade-off between the extent of the embedding space covered and the computational complexity of FaceQAN. We chose $p=5$ since this enables the full use of values on $[0,1]$. The benchmark images are cropped and aligned as specified per the chosen face recognition model. The same procedure is also used for the baselines unless a different preprocessing approach was proposed or included in the paper describing the baseline.  
All experiments are conducted on a desktop PC with an Intel i9-10900KF CPU, $64$ GB of RAM and an Nvidia $3090$ GPU. 

\begin{table}[!t]
    \centering
    \caption{
    Comparison with state-of-the-art - AUC@FMR$1$e-$3$ [$\times 10^{-3}$] ($\downarrow$). 
    \textbf{B} - Best overall,
    \begin{tabular}{c}\cellcolor{blue!10}\end{tabular} - Best in COMP, \begin{tabular}{c}\cellcolor{red!10}\end{tabular} - Best in CLOSE\vspace{-2mm}}
    \resizebox{0.5\textwidth}{!}{%
    \begin{tabular}{| c| c||c|c|c|c|c|c||c|c|c|}
    
         \hline
         \multicolumn{2}{|c||}{\textbf{FR Model}} & \multicolumn{9}{c|}{\textbf{ArcFace -- AUC@FMR1E-3 [$\times 10^{-3}$]}} \\
         \hline
         \multirow{2}{*}{DT$^\dagger$} & \multirow{2}{*}{DR$^\dagger$}& \multicolumn{6}{c||}{Comparison Baselines (COMP)$^\ddagger$} & \multicolumn{3}{c|}{Closely Related (CLOSE)$^\ddagger$} \\ \cline{3-11}
         & &  FQN & SDD & PFE & PCNet & MagFace & LQN & SER & \textbf{FQ(C)$^\dagger$} & \textbf{FQ(S)$^\dagger$} \\
         
         \hline
         
         \parbox[t]{2mm}{\multirow{4}{*}{\rotatebox[origin=c]{90}{LFW}}}
         
            & $10\%$ & $0.49$ & $0.44$ & $0.45$ & $0.4$ & $0.47$ & \cellcolor{blue!10}$\mathbf{0.38}$ & $0.44$ & \cellcolor{red!10}$0.43$ & $0.43$\\
            & $20\%$ & $0.87$ & $0.73$ & $0.82$ & \cellcolor{blue!10}$0.69$ & $0.89$ & $0.7$ & $0.77$ & \cellcolor{red!10}$\mathbf{0.66}$ & $0.66$\\
            & $40\%$ & $1.49$ & $1.37$ & $1.4$ & \cellcolor{blue!10}$1.25$ & $1.43$ & $1.28$ & $1.3$ & \cellcolor{red!10}$\mathbf{1.03}$ & $1.03$\\
            & $80\%$ & $2.52$ & $2.2$ & $2.36$ & \cellcolor{blue!10}$\mathbf{1.73}$ & $2.08$ & $2.34$ & $2.49$ & $1.84$ & \cellcolor{red!10}$1.83$\\
         
          \hline
          
          \parbox[t]{2mm}{\multirow{4}{*}{\rotatebox[origin=c]{90}{CFP-FP}}}
          
            & $10\%$ & $2.07$ & \cellcolor{blue!10}$1.28$ & $1.37$ & $1.3$ & $1.53$ & $1.65$ & $0.92$ & \cellcolor{red!10}$\mathbf{0.81}$ & $0.82$\\
            & $20\%$ & $3.83$ & \cellcolor{blue!10}$1.74$ & $2.01$ & $1.84$ & $2.39$ & $2.75$ & $1.3$ & \cellcolor{red!10}$\mathbf{0.96}$ & $0.98$\\
            & $40\%$ & $6.29$ & \cellcolor{blue!10}$2.15$ & $2.74$ & $2.46$ & $3.14$ & $4.33$ & $1.87$ & \cellcolor{red!10}$\mathbf{1.17}$ & $1.17$\\
            & $80\%$ & $8.75$ & $3.0$ & $3.26$ & \cellcolor{blue!10}$2.95$ & $3.77$ & $5.95$ & $2.67$ & \cellcolor{red!10}$\mathbf{1.7}$ & $1.74$\\
          
          \hline
          
          \parbox[t]{2mm}{\multirow{4}{*}{\rotatebox[origin=c]{90}{XQLFW}}}
          
            & $10\%$ & $55.71$ & $54.99$ & $54.51$ & \cellcolor{blue!10}$53.26$ & $54.44$ & $53.84$ & \cellcolor{red!10}$\mathbf{52.06}$ & $53.19$ & $53.24$\\
            & $20\%$ & $110.68$ & $107.41$ & $104.84$ & \cellcolor{blue!10}$94.59$ & $105.72$ & $97.17$ & \cellcolor{red!10}$\mathbf{90.9}$ & $93.56$ & $93.7$\\
            & $40\%$ & $219.04$ & $188.38$ & $175.9$ & \cellcolor{blue!10}$152.36$ & $203.53$ & $157.45$ & $142.6$ & \cellcolor{red!10}$\mathbf{139.48}$ & $143.3$\\
            & $80\%$ & $422.55$ & $286.99$ & $231.52$ & \cellcolor{blue!10}$199.0$ & $371.23$ & $210.83$ & $187.38$ & \cellcolor{red!10}$\mathbf{175.51}$ & $181.18$\\
          
          \hline
          
          \parbox[t]{2mm}{\multirow{4}{*}{\rotatebox[origin=c]{90}{IJB-C}}}
            & $10\%$ & $3.29$ & $2.64$ & $2.72$ & \cellcolor{blue!10}$2.51$ & $2.7$ & $2.64$ & \cellcolor{red!10}$\mathbf{2.3}$ & $2.37$ & $2.41$\\
            & $20\%$ & $7.14$ & $4.58$ & $4.91$ & \cellcolor{blue!10}$4.4$ & $5.11$ & $4.51$ & $4.14$ & \cellcolor{red!10}$\mathbf{4.14}$ & $4.25$\\
            & $40\%$ & $16.87$ & \cellcolor{blue!10}$7.72$ & $8.7$ & $7.86$ & $9.28$ & $7.94$ & $7.47$ & \cellcolor{red!10}$\mathbf{7.32}$ & $7.49$\\
            & $80\%$ & $42.12$ & $13.56$ & $14.6$ & $13.7$ & $15.36$ & \cellcolor{blue!10}$13.01$ & $13.84$ & \cellcolor{red!10}$\mathbf{12.11}$ & $12.21$\\
          
          \hline\hline

         \multicolumn{2}{|c||}{\textbf{FR Model}} & \multicolumn{9}{c|}{\textbf{CurricularFace -- AUC@FMR1E-3 [$\times 10^{-3}$]}} \\
         \hline
         \multirow{2}{*}{DT$^\dagger$} & \multirow{2}{*}{DR$^\dagger$}& \multicolumn{6}{c||}{Comparison Baselines (COMP)$^\ddagger$} & \multicolumn{3}{c|}{Closely Related (CLOSE)$^\ddagger$} \\ \cline{3-11}
         & &  FQN & SDD & PFE & PCNet & MagFace & LQN & SER & \textbf{FQ(C)$^\dagger$} & \textbf{FQ(S)$^\dagger$} \\
         
         \hline
         
         \parbox[t]{2mm}{\multirow{4}{*}{\rotatebox[origin=c]{90}{LFW}}}
          
            & $10\%$ & \cellcolor{blue!10}$\mathbf{0.32}$ & $0.38$ & $0.45$ & $0.37$ & $0.42$ & $0.38$ & $0.45$ & \cellcolor{red!10}$0.37$ & $0.41$\\
            & $20\%$ & \cellcolor{blue!10}$\mathbf{0.57}$ & $0.65$ & $0.81$ & $0.72$ & $0.78$ & $0.71$ & $0.87$ & \cellcolor{red!10}$0.58$ & $0.76$\\
            & $40\%$ & \cellcolor{blue!10}$1.04$ & $1.25$ & $1.35$ & $1.31$ & $1.36$ & $1.21$ & $1.54$ & \cellcolor{red!10}$\mathbf{0.99}$ & $1.34$\\
            & $80\%$ & \cellcolor{blue!10}$1.9$ & $2.0$ & $2.38$ & $1.93$ & $2.01$ & $2.19$ & $3.0$ & $1.66$ & \cellcolor{red!10}$\mathbf{1.53}$\\
         
          \hline
          
          \parbox[t]{2mm}{\multirow{4}{*}{\rotatebox[origin=c]{90}{CFP-FP}}}
          
            & $10\%$ & $2.2$ & $1.56$ & $1.44$ & \cellcolor{blue!10}$1.37$ & $1.64$ & $1.8$ & $1.19$ & $1.55$ & \cellcolor{red!10}$\mathbf{1.03}$\\
            & $20\%$ & $3.91$ & $2.09$ & $2.04$ & \cellcolor{blue!10}$1.83$ & $2.54$ & $2.85$ & $1.68$ & $2.45$ & \cellcolor{red!10}$\mathbf{1.25}$\\
            & $40\%$ & $6.39$ & $2.5$ & $2.65$ & \cellcolor{blue!10}$2.39$ & $3.3$ & $4.17$ & $2.37$ & $3.96$ & \cellcolor{red!10}$\mathbf{1.55}$\\
            & $80\%$ & $8.7$ & $3.37$ & $3.19$ & \cellcolor{blue!10}$3.0$ & $3.93$ & $5.61$ & $3.16$ & $5.99$ & \cellcolor{red!10}$\mathbf{2.17}$\\
          
          \hline
          
          \parbox[t]{2mm}{\multirow{4}{*}{\rotatebox[origin=c]{90}{XQLFW}}}
          
            & $10\%$ & $74.15$ & $73.89$ & $73.56$ & \cellcolor{blue!10}$\mathbf{67.69}$ & $73.85$ & $70.45$ & $70.22$ & \cellcolor{red!10}$68.48$ & $70.66$\\
            & $20\%$ & $147.75$ & $146.09$ & $142.97$ & \cellcolor{blue!10}$\mathbf{124.55}$ & $146.23$ & $134.75$ & $132.63$ & \cellcolor{red!10}$129.23$ & $136.07$\\
            & $40\%$ & $294.9$ & $269.95$ & $252.79$ & \cellcolor{blue!10}$213.54$ & $282.63$ & $227.54$ & $221.3$ & \cellcolor{red!10}$\mathbf{211.55}$ & $236.28$\\
            & $80\%$ & $579.14$ & $429.54$ & $338.31$ & $293.0$ & $529.29$ & \cellcolor{blue!10}$281.82$ & $291.27$ & \cellcolor{red!10}$\mathbf{272.03}$ & $316.06$\\
          
          \hline
          
          \parbox[t]{2mm}{\multirow{4}{*}{\rotatebox[origin=c]{90}{IJB-C}}}
          
            & $10\%$ & $3.69$ & $2.91$ & $3.01$ & \cellcolor{blue!10}$2.75$ & $3.01$ & $2.91$ & \cellcolor{red!10}$\mathbf{2.53}$ & $2.58$ & $2.62$\\
            & $20\%$ & $7.98$ & $5.04$ & $5.43$ & \cellcolor{blue!10}$4.87$ & $5.65$ & $4.93$ & \cellcolor{red!10}$\mathbf{4.49}$ & $4.54$ & $4.63$\\
            & $40\%$ & $18.86$ & \cellcolor{blue!10}$8.37$ & $9.42$ & $8.58$ & $9.92$ & $8.49$ & $7.96$ & \cellcolor{red!10}$\mathbf{7.95}$ & $8.05$\\
            & $80\%$ & $47.14$ & $14.21$ & $15.36$ & $14.51$ & $16.1$ & \cellcolor{blue!10}$13.61$ & $14.46$ & $12.86$ & \cellcolor{red!10}$\mathbf{12.74}$\\
          
          \hline\hline

          \multicolumn{2}{|c||}{\textbf{FR Model}} & \multicolumn{9}{c|}{\textbf{ElasticFace -- AUC@FMR1E-3 [$\times 10^{-3}$]}} \\
         \hline
         \multirow{2}{*}{DT$^\dagger$} & \multirow{2}{*}{DR$^\dagger$}& \multicolumn{6}{c||}{Comparison Baselines (COMP)$^\ddagger$} & \multicolumn{3}{c|}{Closely Related (CLOSE)$^\ddagger$} \\ \cline{3-11}
         & &  FQN & SDD & PFE & PCNet & MagFace & LQN & SER & \textbf{FQ(C)$^\dagger$} & \textbf{FQ(S)$^\dagger$} \\
         
         \hline
         
         \parbox[t]{2mm}{\multirow{4}{*}{\rotatebox[origin=c]{90}{LFW}}}
            & $10\%$ & \cellcolor{blue!10}$\mathbf{0.57}$ & $0.68$ & $0.66$ & $0.72$ & $0.69$ & $0.59$ & $0.74$ & \cellcolor{red!10}$0.59$ & $0.59$\\
            & $20\%$ & \cellcolor{blue!10}$0.97$ & $1.14$ & $1.17$ & $1.32$ & $1.18$ & $1.05$ & $1.23$ & \cellcolor{red!10}$\mathbf{0.85}$ & $0.95$\\
            & $40\%$ & \cellcolor{blue!10}$1.68$ & $1.94$ & $1.77$ & $2.24$ & $1.86$ & $1.69$ & $2.0$ & \cellcolor{red!10}$\mathbf{1.24}$ & $1.42$\\
            & $80\%$ & $2.94$ & $2.97$ & $2.79$ & $3.02$ & \cellcolor{blue!10}$2.53$ & $2.69$ & $3.57$ & $1.88$ & \cellcolor{red!10}$\mathbf{1.61}$\\
         
          \hline
          
          \parbox[t]{2mm}{\multirow{4}{*}{\rotatebox[origin=c]{90}{CFP-FP}}}
          
            & $10\%$ & $2.42$ & $1.78$ & \cellcolor{blue!10}$1.69$ & $1.71$ & $1.88$ & $2.05$ & $1.42$ & \cellcolor{red!10}$\mathbf{1.16}$ & $1.22$\\
            & $20\%$ & $4.25$ & $2.44$ & $2.43$ & \cellcolor{blue!10}$2.34$ & $2.95$ & $3.26$ & $2.08$ & \cellcolor{red!10}$\mathbf{1.45}$ & $1.55$\\
            & $40\%$ & $6.82$ & \cellcolor{blue!10}$2.95$ & $3.17$ & $2.99$ & $3.84$ & $4.76$ & $2.86$ & $1.84$ & \cellcolor{red!10}$\mathbf{1.8}$\\
            & $80\%$ & $9.04$ & $3.87$ & $3.71$ & \cellcolor{blue!10}$3.49$ & $4.49$ & $6.26$ & $3.68$ & $2.3$ & \cellcolor{red!10}$\mathbf{2.28}$\\
          
          \hline
          
          \parbox[t]{2mm}{\multirow{4}{*}{\rotatebox[origin=c]{90}{XQLFW}}}
          
            & $10\%$ & $75.04$ & $74.67$ & $74.54$ & $73.4$ & $75.02$ & \cellcolor{blue!10}$73.23$ & \cellcolor{red!10}$\mathbf{71.9}$ & $74.21$ & $74.36$\\
            & $20\%$ & $149.38$ & $147.68$ & $146.64$ & \cellcolor{blue!10}$131.99$ & $148.6$ & $135.74$ & \cellcolor{red!10}$\mathbf{130.25}$ & $134.91$ & $144.68$\\
            & $40\%$ & $295.97$ & $282.37$ & $280.81$ & \cellcolor{blue!10}$\mathbf{219.83}$ & $291.61$ & $226.24$ & $225.59$ & \cellcolor{red!10}$219.84$ & $247.56$\\
            & $80\%$ & $577.87$ & $460.28$ & $393.29$ & $301.86$ & $534.96$ & \cellcolor{blue!10}$297.76$ & $312.28$ & \cellcolor{red!10}$\mathbf{288.4}$ & $320.85$\\
          
          \hline
          
          \parbox[t]{2mm}{\multirow{4}{*}{\rotatebox[origin=c]{90}{IJB-C}}}
        
            & $10\%$ & $3.65$ & $2.79$ & $2.79$ & \cellcolor{blue!10}$2.66$ & $2.87$ & $2.79$ & \cellcolor{red!10}$\mathbf{2.4}$ & $2.47$ & $2.51$\\
            & $20\%$ & $7.96$ & $4.78$ & $5.03$ & \cellcolor{blue!10}$4.6$ & $5.39$ & $4.68$ & $4.3$ & \cellcolor{red!10}$\mathbf{4.28}$ & $4.32$\\
            & $40\%$ & $18.91$ & \cellcolor{blue!10}$8.0$ & $8.92$ & $8.14$ & $9.62$ & $8.16$ & $7.83$ & \cellcolor{red!10}$\mathbf{7.46}$ & $7.48$\\
            & $80\%$ & $47.41$ & $13.99$ & $14.98$ & $14.04$ & $15.86$ & \cellcolor{blue!10}$13.37$ & $14.26$ & $12.28$ & \cellcolor{red!10}$\mathbf{12.13}$\\
          
          \hline

         \multicolumn{11}{l}{\footnotesize $^\dagger$DT -- Dataset; DR -- Drop Rate (or Ratio of unconsidered images). (C) -- cross-model; (S) -- model-specific}\\
         \multicolumn{11}{l}{\footnotesize $^\ddagger$FQN -- FaceQNet; SDD -- SDD- FIQA; LQN -- LightQNet; SER -- SER-FIQ, FQ - FaceQAN (ours)}\vspace{-8mm}\\
    \end{tabular}
    }
    \label{tab:auc_res}
\end{table}
\begin{table}[!t]
    \centering
    \caption{Analysis of the time complexity over CFP-FP using CosFace}\vspace{-2mm}
    \resizebox{0.99\columnwidth}{!}{%
    \begin{tabular}{|c c|c| c| c| c| c| c| c|}
    \hline
        \multicolumn{2}{|l|}{\textbf{Complexity}} &  \textbf{FaceQnet} & \textbf{SDD-FIQA} & \textbf{PFE} & \textbf{PCNet} & \textbf{MagFace} & \textbf{LightQnet} & \textbf{SER-FIQ} \\
        \hline
        \multirow{2}{*}{t [s]}& $\mu$ & $0.0432$ & $0.0006$  & $ 0.0493 $ & 
        $ 0.0175 $ & $0.0011 $ & $0.0535 $ & $0.1125 $\\
        & $\sigma$ & $0.0026$ & $0.0004$  & $0.0275$ & 
        $ 0.0004$ & $0.0004$ & $0.0468$ & $0.0418$\\

    \hline
    \end{tabular}
    
    }
    \vspace{-1mm}
    \resizebox{0.99\columnwidth}{!}{%
    \begin{tabular}{|c c | c|c|c|c|c|}
    \hline
       \multicolumn{2}{|l|}{\multirow{2}{*}{\textbf{Complexity}}} &  \multicolumn{5}{c|}{\textbf{FaceQAN (ours)}}\\ \cline{3-7}
        & & $k=2$ & $k=5$ & $^\dagger k=10$ & $k=50$ & $k=100$\\
    \hline
        \multicolumn{2}{|l|}{t [s] $(\mu \pm \sigma$)} & $0.21 \pm 0.019$ & $0.23 \pm 0.018$ & $0.30 \pm 0.006$ & $1.00 \pm 0.027$ & $1.87 \pm 0.031$\\
    \hline
    \multicolumn{7}{l}{$\dagger$ Configuration used in experiments}\vspace{-4mm}
    
    \end{tabular}
    
    }
    
    \label{tab:complex}
\end{table}


\subsection{Comparison with the State-Of-The-Art}

The ERC plots for all combinations of FR models and datasets are shown in Fig.~\ref{fig:erc_res} and the calculated AUC scores  
over the ERC plots (multiplied by $10^3$ for readability) in Table~\ref{tab:auc_res}. Below we analyze these results from two aspects: $(i)$ in comparison to all considered (supervised) baselines (COMP; see Table~\ref{tab:auc_res}), and $(ii)$ in comparison to the closely-related SER-FIQ, which does not rely on training (CLOSE). 

\begin{table}[t!]
    \begin{center}
    \caption{AUC [$\times 10^{-3}$] scores  ($\downarrow$) of the ablation study 
    }\vspace{-2mm}
\resizebox{0.8\columnwidth}{!}{%
\centering
    \begin{tabular}{|c|l|l|c|c|c|c|}
        \hline
        \multirow{ 2}{*}{\textbf{FR}} & \multirow{ 2}{*}{\textbf{Dataset}} & \multirow{ 2}{*}{\textbf{FIQA Model}} & \multicolumn{4}{c|}{\textbf{Image Drop Rate}}\\ \cline{4-7}
        & & & $10\%$ & $20\%$ & $40\%$ & $80\%$  \\
        \hline
        \hline
        \parbox[t]{2mm}{\multirow{4}{*}{\rotatebox[origin=c]{90}{\scriptsize \textbf{ArcFace}}}} 
        & \multirow{2}{*}{LFW} &
        FaceQAN         & $\mathbf{0.43}$ & $\mathbf{0.66}$ & $\mathbf{1.03}$ & $\mathbf{1.83}$ \\
        & & w/o Symm. Est.  & $0.53$ &  $0.99$ & $1.5$  & $2.5$ \\
        \cline{2-7}
        & \multirow{2}{*}{CFP-FP} &
          FaceQAN         & $\mathbf{0.82}$ & $\mathbf{0.98}$ & $\mathbf{1.17}$ & $\mathbf{1.74}$ \\ 
        & & w/o Symm. Est.  & $0.89$ & $1.25$ & $1.56$ & $2.21$ \\
        \hline
        \hline
        \parbox[t]{2mm}{\multirow{4}{*}{\rotatebox[origin=c]{90}{\scriptsize \textbf{Cu. Face$^\dagger$}}}} 
        & \multirow{2}{*}{LFW} &
        FaceQAN         & $0.41$ & $0.76$ & $1.34$ & $\mathbf{1.53}$ \\
        & & w/o Symm. Est.  & $\mathbf{0.35}$ & $\mathbf{0.67}$ & $\mathbf{1.31}$  & $1.63$ \\
        \cline{2-7}
        & \multirow{2}{*}{CFP-FP} &
          FaceQAN         & $\mathbf{1.03}$ & $\mathbf{1.25}$ & $\mathbf{1.55}$ & $\mathbf{2.17}$ \\ 
        & & w/o Symm. Est.  & $1.06$ & $1.4$ & $1.75$ & $2.48$ \\
        \hline
        \hline
        \parbox[t]{2mm}{\multirow{4}{*}{\rotatebox[origin=c]{90}{\scriptsize \textbf{ElasticFace}}}} 
        & \multirow{2}{*}{LFW} &
        FaceQAN         & $\mathbf{0.59}$ & $\mathbf{0.95}$ & $\mathbf{1.42}$ & $\mathbf{1.61}$ \\
        & & w/o Symm. Est.  & $0.63$ & $1.05$ & $1.65$  & $2.03$ \\
        \cline{2-7}
        & \multirow{2}{*}{CFP-FP} &
          FaceQAN         & $\mathbf{1.22}$ & $\mathbf{1.55}$ & $\mathbf{1.8}$ & $\mathbf{2.28}$ \\ 
        & & w/o Symm. Est.  & $1.38$ & $1.89$ & $2.62$ & $3.32$ \\
        \hline

        \multicolumn{7}{l}{$\dagger$~\footnotesize Cu.Face -- CurricularFace}    \vspace{-8mm}    
        
    \end{tabular}
    }
    
    \label{tab:my_flips}
    \end{center}

\end{table}

\textbf{Baseline Comparisons (COMP).} 
FaceQAN is the most convincing approach when compared to the state-of-the-art methods from the COMP group on all four dataset when considering the ArcFace model despite not relying on supervised training or (pseudo) reference labels. 
Similar observations can also be made for the CurricularFace and ElasticFace models, where FaceQAN again yields highly competitive results on all four datasets in the cross-model (C) and model-specific setting (S). The only exeption is the cross-model result with CurricularFace on the CFP-FP dataset, where FaceQAN is less convincing. 
It is also interesting to observe that the relative ranking of the competing FIQA techniques changes with different FR models. FaceQAN, on the other hand, is among the top performers with all three considered FR models. 

\textbf{Comparison to SER-FIQ (CLOSE).} SER-FIQ and FaceQAN are the only models in our experiments that require no supervision and estimate face quality solely based on the input image and a FR model. As can be seen from Table~\ref{tab:auc_res}, the cross-model version of FaceQAN(C) is overall the top performer, suggesting that the adversarial examples generated with the CosFace model are particularly informative for quality estimation and therefore often outperform FaceQAN in the model-specific setting. SER-FIQ is most competitive on the XQLFW dataset. However, it needs to be noted that XQLFW was designed by optimizing SER-FIQ quality scores and is, therefore, highly biased towards this FIQA approach~\cite{xqlfw}. Convincing results are also achieved with SER-FIQ on the large-scale IJB-C dataset at the lowest drop rate, where the performance is comparable to FaceQAN(C). 

\textbf{Time Complexity.} In Table \ref{tab:complex} we analyze the time complexity of FaceQAN in comparison with the considered baselines and as a function of the number of generated adversarial examples $k$. The analysis is done over the CFP-FP dataset and with the CosFace model. As expected, the time complexity of FaceQAN increases with an increase in $k$ and is overall somewhat higher than that of the competing solutions due to the nature of the BIM approach selected for our implementation.

\subsection{Ablation Study}
To demonstrate the impact of the symmetry estimation step on the performance of FaceQAN(S), we perform an ablation study on the LFW and CFP-FP datasets using the three selected FR models (ArcFace, CurricularFace, ElasticFace) with and without (w/o) the symmetry scores $s_f$ from Eq.~\eqref{eq:pose}. As can be seen from the results in Table \ref{tab:my_flips}, removing the step from the quality assessment process has a detrimental effect on the AUC scores, which consistently increase in comparison to the full FaceQAN model across all tested combinations, except when using CurricularFace on the LFW dataset. 

\subsection{Qualitative Evaluation} 
\begin{figure}[t]
    \centering
    \includegraphics[width=0.92\columnwidth, trim = 0mm 0mm 0mm 0mm,clip]{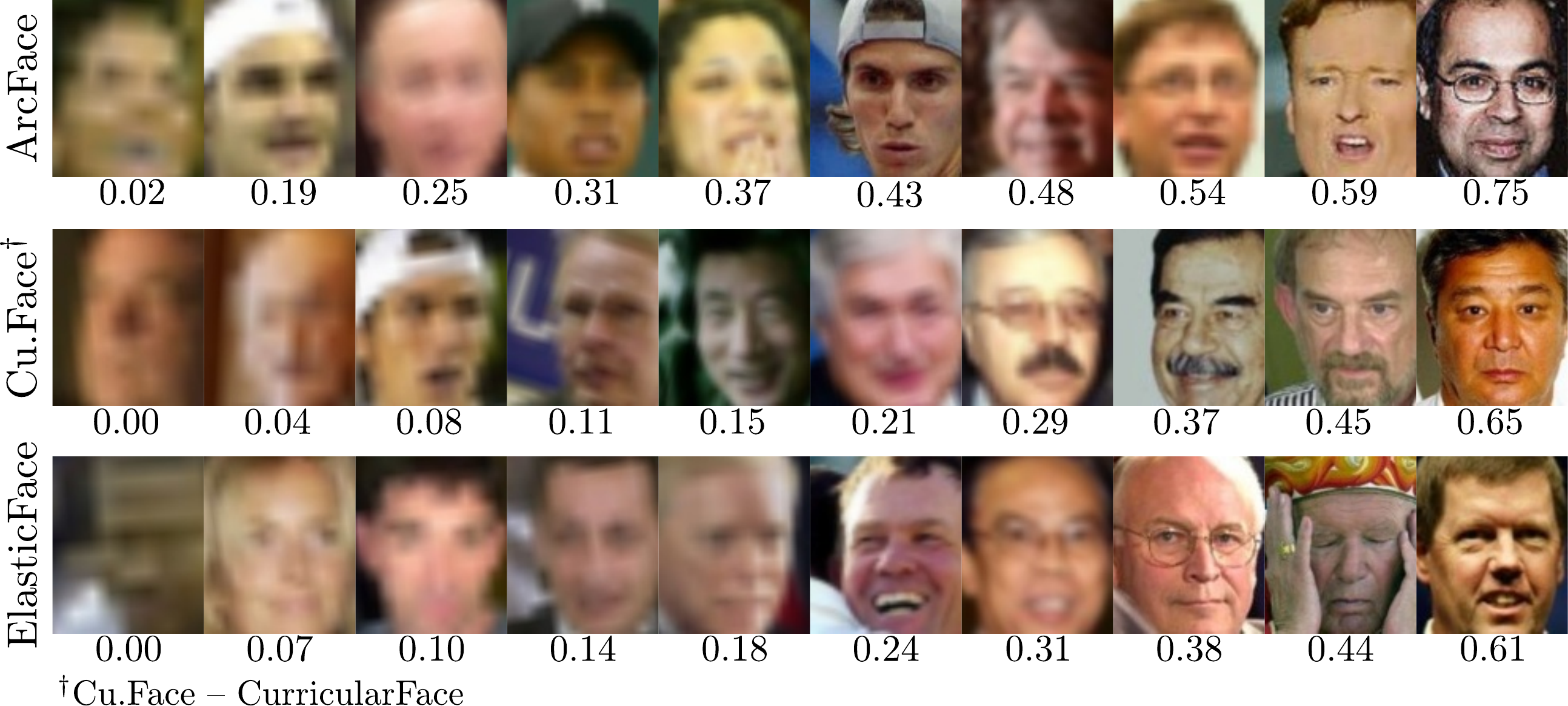}\vspace{-2mm}
    \caption{\textbf{Quality-based ranking with FaceQAN.} Note how the ordering of the XQLFW sample images  based on the generated quality scores (shown below the images) corresponds to the perceived face quality. 
    Zoom-in for details.}\vspace{-3mm}
    \label{fig:vis_il}
\end{figure}

\textbf{Image Ranking.} In Fig. \ref{fig:vis_il} we show example images from XQLFW ordered according to the computed quality scores. As can be seen, the images follow a reasonable order in terms of perceived quality for all three FR models. Because the ordering is meant to reflect the utility of the samples for face recognition, it is interesting to see how FaceQAN(S) favors blurry frontal images in certain settings over crisp, but less frontal (and w/o neutral expressions)  samples. 

\begin{figure}[h]
    \centering
    \includegraphics[width=0.97\columnwidth, trim = 0mm 0mm 0mm 0 mm,clip]{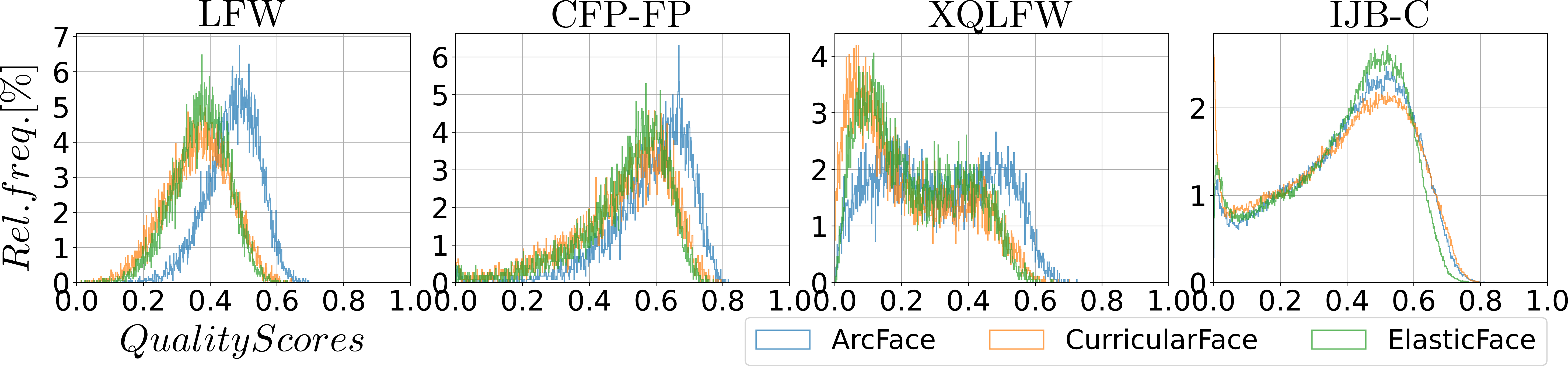}
    \caption{\textbf{Quality-score distribution generated by FaceQAN.} Distributions are shown for (from left to right): LFW,  CFP-FP, XQLFW, and IJB-C. The three FR models produce distributions on different scales but of similar shape.}\vspace{-2mm}
    \label{fig:vis_dist}
\end{figure}

\textbf{Quality-Score Distribution.}  Fig.~\ref{fig:vis_dist} shows that the quality-score distributions generated for the ArcFace, CurricularFace and ElasticFace models exhibit a relatively similar shape, but have a somewhat different support on some datasets. 
Overall, however, all models generate reasonably consistent score distributions  (also given the results in Fig.~\ref{fig:vis_il}) that well capture the quality/utility of the input data.  

\section{Conclusion}\label{conclusion}

We presented a novel approach to unsupervised face image quality assessment, called FaceQAN. 
The proposed approach elegantly avoids the quality-label generation and learning process used by many state-of-the-art FIQA techniques by harnessing adversarial noise, which can be generated for any modern deep learning model \cite{akhtar2021advances}. By comparing the embeddings of adversarial examples and the original input sample, FaceQAN is able to calculate a quality score that is an excellent predictor of the sample's utility for face recognition. Extensive experiments with several baselines, datasets and FR models have shown that FaceQAN achieves highly competitive results, while being based on minimal assumptions. As part of our future work, we plan to explore the use of adversarial examples in supervised settings and the predictive power of adversarial noise with (pseudo) reference quality labels. 

\section*{Acknowledgments}

Supported by the ARRS Research Program P2--0250 (B) as well as the ARRS Junior Researcher Program.






\bibliographystyle{IEEEtran}
\bibliography{egbib}

\begin{thebibliography}{10}
\providecommand{\url}[1]{#1}
\csname url@samestyle\endcsname
\providecommand{\newblock}{\relax}
\providecommand{\bibinfo}[2]{#2}
\providecommand{\BIBentrySTDinterwordspacing}{\spaceskip=0pt\relax}
\providecommand{\BIBentryALTinterwordstretchfactor}{4}
\providecommand{\BIBentryALTinterwordspacing}{\spaceskip=\fontdimen2\font plus
\BIBentryALTinterwordstretchfactor\fontdimen3\font minus
  \fontdimen4\font\relax}
\providecommand{\BIBforeignlanguage}[2]{{%
\expandafter\ifx\csname l@#1\endcsname\relax
\typeout{** WARNING: IEEEtran.bst: No hyphenation pattern has been}%
\typeout{** loaded for the language `#1'. Using the pattern for}%
\typeout{** the default language instead.}%
\else
\language=\csname l@#1\endcsname
\fi
#2}}
\providecommand{\BIBdecl}{\relax}
\BIBdecl

\bibitem{grm2018strengths}
K.~Grm, V.~{\v{S}}truc, A.~Artiges, M.~Caron, and H.~K. Ekenel, ``Strengths and
  weaknesses of deep learning models for face recognition against image
  degradations,'' \emph{IET Biometrics}, vol.~7, no.~1, pp. 81--89, 2018.

\bibitem{grm2018deep}
K.~Grm and V.~Struc, ``Deep face recognition for surveillance applications,''
  \emph{IEEE Intelligent Systems}, vol.~33, no.~3, pp. 46--50, 2018.

\bibitem{meden2021privacy}
B.~Meden, P.~Rot, P.~Terh{\"o}rst, N.~Damer, A.~Kuijper, W.~J. Scheirer,
  A.~Ross, P.~Peer, and V.~{\v{S}}truc, ``Privacy--enhancing face biometrics: A
  comprehensive survey,'' \emph{IEEE Transactions on Information Forensics and
  Security}, 2021.

\bibitem{wang2021deep}
M.~Wang and W.~Deng, ``Deep face recognition: A survey,''
  \emph{Neurocomputing}, vol. 429, pp. 215--244, 2021.

\bibitem{MFR_IJCB2021}
F.~Boutros, N.~Damer, J.~N. Kolf, K.~Raja, F.~Kirchbuchner, R.~Ramachandra,
  A.~Kuijper, P.~Fang, C.~Zhang, F.~Wang, D.~Montero, N.~Aginako, B.~Sierra,
  M.~Nieto, M.~E. Erakin, U.~Demir, H.~K. Ekenel, A.~Kataoka, K.~Ichikawa,
  S.~Kubo, J.~Zhang, M.~He, D.~Han, S.~Shan, K.~Grm, V.~Štruc, S.~Seneviratne,
  N.~Kasthuriarachchi, S.~Rasnayaka, P.~C. Neto, A.~F. Sequeira, J.~R. Pinto,
  M.~Saffari, and J.~S. Cardoso, ``{MFR 2021: Masked Face Recognition
  Competition},'' in \emph{Proceedings of the IEEE International Joint
  Conference on Biometrics (IJCB)}, 2021.

\bibitem{grm2019face}
K.~Grm, W.~J. Scheirer, and V.~{\v{S}}truc, ``Face hallucination using cascaded
  super-resolution and identity priors,'' \emph{IEEE Transactions on Image
  Processing}, vol.~29, pp. 2150--2165, 2020.

\bibitem{surveypaper}
T.~Schlett, C.~Rathgeb, O.~Henniger, J.~Galbally, J.~Fierrez, and C.~Busch,
  ``Face image quality assessment: A literature survey,'' \emph{ACM Computing
  Surveys}, 2022.

\bibitem{isoiec}
{ISO/IEC JTC 1/SC 37 Biometrics}, ``{Information Technology} - {Biometric
  Sample Quality} - {Part 1: Framework},'' International Organization for
  Standardization, Standard ISO/IEC 29794-1:2016, 2016.

\bibitem{pcnet}
W.~Xie, J.~Byrne, and A.~Zisserman, ``Inducing predictive uncertainty
  estimation for face verification,'' in \emph{British Machine Vision
  Conference (BMVC)}, 2020.

\bibitem{lightqnet}
K.~Chen, T.~Yi, and Q.~Lv, ``Lightqnet: Lightweight deep face quality
  assessment for risk-controlled face recognition,'' \emph{IEEE Signal
  Processing Letters}, vol.~28, pp. 1878--1882, 2021.

\bibitem{faceqnet}
J.~Hernandez-Ortega, J.~Galbally, J.~Fierrez, R.~Haraksim, and L.~Beslay,
  ``Faceqnet: Quality assessment for face recognition based on deep learning,''
  in \emph{Proceedings of the IEEE International Conference on Biometrics
  (ICB)}, 2019, pp. 1--8.

\bibitem{pfe}
Y.~Shi and A.~K. Jain, ``Probabilistic face embeddings,'' in \emph{Proceedings
  of the IEEE International Conference on Computer Vision (ICCV)}, 2019, pp.
  6902--6911.

\bibitem{magface}
Q.~Meng, S.~Zhao, Z.~Huang, and F.~Zhou, ``Magface: A universal representation
  for face recognition and quality assessment,'' in \emph{CVF/IEEE Conference
  on Computer Vision and Pattern Recognition (CVPR)}, 2021, pp.
  14\,225--14\,234.

\bibitem{serfiq}
P.~Terhorst, J.~N. Kolf, N.~Damer, F.~Kirchbuchner, and A.~Kuijper, ``{SER-FIQ:
  Unsupervised Estimation of Face Image Quality Based on Stochastic Embedding
  Robustness},'' in \emph{CVF/IEEE Conference on Computer Vision and Pattern
  Recognition (CVPR)}, 2020, pp. 5651--5660.

\bibitem{akhtar2021advances}
N.~Akhtar, A.~Mian, N.~Kardan, and M.~Shah, ``Advances in adversarial attacks
  and defenses in computer vision: A survey,'' \emph{IEEE Access}, vol.~9, pp.
  155\,161--155\,196, 2021.

\bibitem{sddfiqa}
O.~Fu-Zhao, X.~Chen, R.~Zhang, Y.~Huang, S.~Li, J.~Li, Y.~Li, L.~Cao, and
  W.~Yuan-Gen, ``{SDD-FIQA: Unsupervised Face Image Quality Assessment with
  Similarity Distribution Distance},'' in \emph{CVF/IEEE Conference on Computer
  Vision and Pattern Recognition (CVPR)}, 2021, pp. 7670--7679.

\bibitem{mix1}
P.~Wasnik, R.~Ramachandra, K.~Raja, and C.~Busch, ``An empirical evaluation of
  deep architectures on generalization of smartphone-based face image quality
  assessment,'' in \emph{IEEE International Conference on Biometrics Theory,
  Applications and Systems (BTAS)}, 2018, pp. 1--7.

\bibitem{mix2}
L.~Best-Rowden and A.~K. Jain, ``Learning face image quality from human
  assessments,'' \emph{IEEE Transactions on Information forensics and
  security}, vol.~13, no.~12, pp. 3064--3077, 2018.

\bibitem{vggface2}
Q.~Cao, L.~Shen, W.~Xie, O.~M. Parkhi, and A.~Zisserman, ``Vggface2: A dataset
  for recognising faces across pose and age,'' in \emph{IEEE International
  Conference on Automatic Face \& Gesture Recognition (FG)}, 2018, pp. 67--74.

\bibitem{biolab-icao}
D.~Maltoni, A.~Franco, M.~Ferrara, D.~Maio, and A.~Nardelli, ``Biolab-icao: A
  new benchmark to evaluate applications assessing face image compliance to
  iso/iec 19794-5 standard,'' in \emph{IEEE International Conference on Image
  Processing (ICIP)}, 2009, pp. 41--44.

\bibitem{isoiec2}
{ISO/IEC JTC 1/SC 37 Biometrics}, ``{Information technology} - {Biometric Data
  Initerchange Formats} - {Part 5: Face Image Data},'' International
  Organization for Standardization, Standard ISO/IEC 19794-5:2011, 2011.

\bibitem{lfw}
G.~B. Huang, M.~Ramesh, T.~Berg, and E.~Learned-Miller, ``Labeled faces in the
  wild: A database for studying face recognition in unconstrained
  environments,'' University of Massachusetts, Amherst, Tech. Rep. 07-49,
  October 2007.

\bibitem{fgsm}
I.~J. Goodfellow, J.~Shlens, and C.~Szegedy, ``Explaining and harnessing
  adversarial examples,'' \emph{arXiv preprint arXiv:1412.6572}, 2014.

\bibitem{madry2017towards}
A.~Madry, A.~Makelov, L.~Schmidt, D.~Tsipras, and A.~Vladu, ``Towards deep
  learning models resistant to adversarial attacks,'' in \emph{International
  Conference on Learning Representations (ICLR)}, 2018.

\bibitem{carlini2017towards}
N.~Carlini and D.~Wagner, ``Towards evaluating the robustness of neural
  networks,'' in \emph{IEEE Symposium on Security and Privacy (SSP)}, 2017, pp.
  39--57.

\bibitem{fgsmextend}
H.~Wang, S.~Wang, Z.~Jin, Y.~Wang, C.~Chen, and T.~Massimo, ``Similarity-based
  gray-box adversarial attack against deep face recognition,'' in \emph{IEEE
  International Conference on Automatic Face \& Gesture Recognition (FG)},
  2021.

\bibitem{bim}
A.~Kurakin, I.~J. Goodfellow, and S.~Bengio, ``Adversarial examples in the
  physical world,'' in \emph{Artificial intelligence safety and
  security}.\hskip 1em plus 0.5em minus 0.4em\relax Chapman and Hall/CRC, 2018,
  pp. 99--112.

\bibitem{vstruc2010complete}
V.~{\v{S}}truc and N.~Pave{\v{s}}i{\'c}, ``The complete gabor-fisher classifier
  for robust face recognition,'' \emph{EURASIP Journal on Advances in Signal
  Processing}, vol. 2010, pp. 1--26, 2010.

\bibitem{zhao2018towards}
J.~Zhao, Y.~Cheng, Y.~Xu, L.~Xiong, J.~Li, F.~Zhao, K.~Jayashree, S.~Pranata,
  S.~Shen, J.~Xing \emph{et~al.}, ``Towards pose invariant face recognition in
  the wild,'' in \emph{CVF/IEEE Conference on Computer Vision and Pattern
  Recognition (CVPR)}, 2018, pp. 2207--2216.

\bibitem{banerjee2018frontalize}
S.~Banerjee, J.~Brogan, J.~Krizaj, A.~Bharati, B.~R. Webster, V.~Struc, P.~J.
  Flynn, and W.~J. Scheirer, ``To frontalize or not to frontalize: Do we really
  need elaborate pre-processing to improve face recognition?'' in \emph{2018
  IEEE Winter Conference on Applications of Computer Vision (WACV)}, 2018, pp.
  20--29.

\bibitem{cfp-fp}
S.~Sengupta, J.~C. Cheng, C.~D. Castillo, V.~M. Patel, R.~Chellappa, and D.~W.
  Jacobs, ``Frontal to profile face verification in the wild,'' in \emph{IEEE
  Winter Conference on Applications of Computer Vision (WACV)}, 2016.

\bibitem{xqlfw}
M.~Knoche, S.~Hormann, and G.~Rigoll, ``Cross-quality lfw: A database for
  analyzing cross-resolution image face recognition in unconstrained
  environments,'' in \emph{IEEE International Conference on Automatic Face and
  Gesture Recognition (FG)}, 2021, pp. 1--5.

\bibitem{ijbc}
B.~Maze, J.~Adams, J.~A. Duncan, N.~Kalka, T.~Miller, C.~Otto, A.~K. Jain,
  W.~T. Niggel, J.~Anderson, J.~Cheney \emph{et~al.}, ``{IARPA Janus
  Benchmark-C: Face dataset and protocol},'' in \emph{International Conference
  on Biometrics (ICB)}, 2018, pp. 158--165.

\bibitem{boutros2021cr}
F.~Boutros, M.~Fang, M.~Klemt, B.~Fu, and N.~Damer, ``{CR-FIQA: Face Image
  Quality Assessment by Learning Sample Relative Classifiability},''
  \emph{arXiv preprint arXiv:2112.06592}, 2021.

\bibitem{arcface}
J.~Deng, J.~Guo, N.~Xue, and S.~Zafeiriou, ``{Arcface: Additive Angular Margin
  Loss for Deep Face Recognition},'' in \emph{CVF/IEEE Conference on Computer
  Vision and Pattern Recognition (CVPR)}, 2019, pp. 4690--4699.

\bibitem{curricularface}
Y.~Huang, Y.~Wang, Y.~Tai, X.~Liu, P.~Shen, S.~Li, J.~Li, and F.~Huang,
  ``{CurricularFace: Adaptive curriculum learning loss for deep face
  recognition},'' in \emph{CVF/IEEE Conference on Computer Vision and Pattern
  Recognition (CVPR)}, 2020, pp. 5901--5910.

\bibitem{elasticface}
F.~Boutros, N.~Damer, F.~Kirchbuchner, and A.~Kuijper, ``Elasticface: Elastic
  margin loss for deep face recognition,'' 2021.

\bibitem{cosface}
H.~Wang, Y.~Wang, Z.~Zhou, X.~Ji, D.~Gong, J.~Zhou, Z.~Li, and W.~Liu,
  ``{CosFace: Large margin cosine loss for deep face recognition},'' in
  \emph{CVF/IEEE Conference on Computer Vision and Pattern Recognition (CVPR)},
  2018, pp. 5265--5274.

\bibitem{an2021partial}
X.~An, X.~Zhu, Y.~Gao, Y.~Xiao, Y.~Zhao, Z.~Feng, L.~Wu, B.~Qin, M.~Zhang,
  D.~Zhang \emph{et~al.}, ``Partial {FC}: Training 10 million identities on a
  single machine,'' in \emph{IEEE/CVF International Conference on Computer
  Vision (ICCV)}, 2021, pp. 1445--1449.

\end{thebibliography}
%



\end{document}